\documentclass[runningheads]{llncs}
\usepackage[table]{xcolor}
\usepackage[T1]{fontenc}
\usepackage{graphicx,verbatim}
\usepackage{amsmath}
\usepackage{amssymb}
\usepackage{float}
\usepackage{booktabs}
\begin{document}
\title{Lesion-Aware Visual-Language Fusion for Automated Image Captioning of Ulcerative Colitis Endoscopic Examinations}
\titlerunning{Lesion-Aware Image Captioning in UC}

\author{
    Alexis Iván López Escamilla\inst{1} \and
    Gilberto Ochoa\inst{1} \and
    Sharib Ali\inst{2}
}

\authorrunning{López Escamilla et al.}

\institute{
    Monterrey Institute of Technology and Higher Education, Mexico \\
    \email{a01166951@tec.mx, gilberto.ochoa@tec.mx}
    \and
    University of Leeds, United Kingdom \\
    \email{s.ali@leeds.ac.uk}
}

\maketitle

\begin{abstract}
We present a lesion-aware image captioning framework for ulcerative colitis (UC), integrating ResNet embeddings, Grad-CAM heatmaps, and CBAM-enhanced attention with a T5 decoder. Clinical metadata—including MES scores, bleeding, and vascular patterns—are incorporated as natural language prompts to guide caption generation. The resulting system produces structured, interpretable, and diagnostically aligned descriptions. Compared to previous approaches, our method improves both captioning quality and MES classification accuracy, offering a clinically meaningful tool for endoscopic reporting.

% Authors must provide keywords and are not allowed to remove this Keyword section.

\end{abstract}
\section{Introduction}

Automated image captioning has achieved strong results in medical imaging, particularly in radiology, using CNNs and transformer-based decoders to produce fluent reports~\cite{vinyals2015show,xu2015show,chen2020generating}. However, in ulcerative colitis (UC), where lesions are subtle and localized, standard models struggle to capture clinically relevant details.

Efforts to bridge the image-text gap via multimodal fusion and memory-driven transformers~\cite{boag2020baselines,li2022medical,huang2020biomedical} improve alignment but often lack explicit lesion grounding. This limits interpretability, a key requirement in clinical practice. The need for explainable models is increasingly emphasized in visual question answering (VQA) and diagnostic captioning~\cite{abacha2019vqa,liu2019clinically}.

While Grad-CAM~\cite{selvaraju2017grad} offers spatial explanations, it is typically used post hoc, limiting its training utility. Recent architectures like HRFormer~\cite{yuan2022hrformer} and hierarchical fusion networks~\cite{zhang2021hierarchical} provide finer attention, but their use in endoscopy remains rare.

We introduce a lesion-aware captioning framework for UC, integrating Grad-CAM maps directly into a ResNet encoder to emphasize pathological regions. CBAM enhances feature saliency, and a T5 decoder generates structured reports. Clinical metadata (MES scores, bleeding, vascular patterns) are embedded as natural language prompts, improving contextual accuracy and report consistency.

\begin{figure}[b!]
    \centering
    \includegraphics[width=1\linewidth]{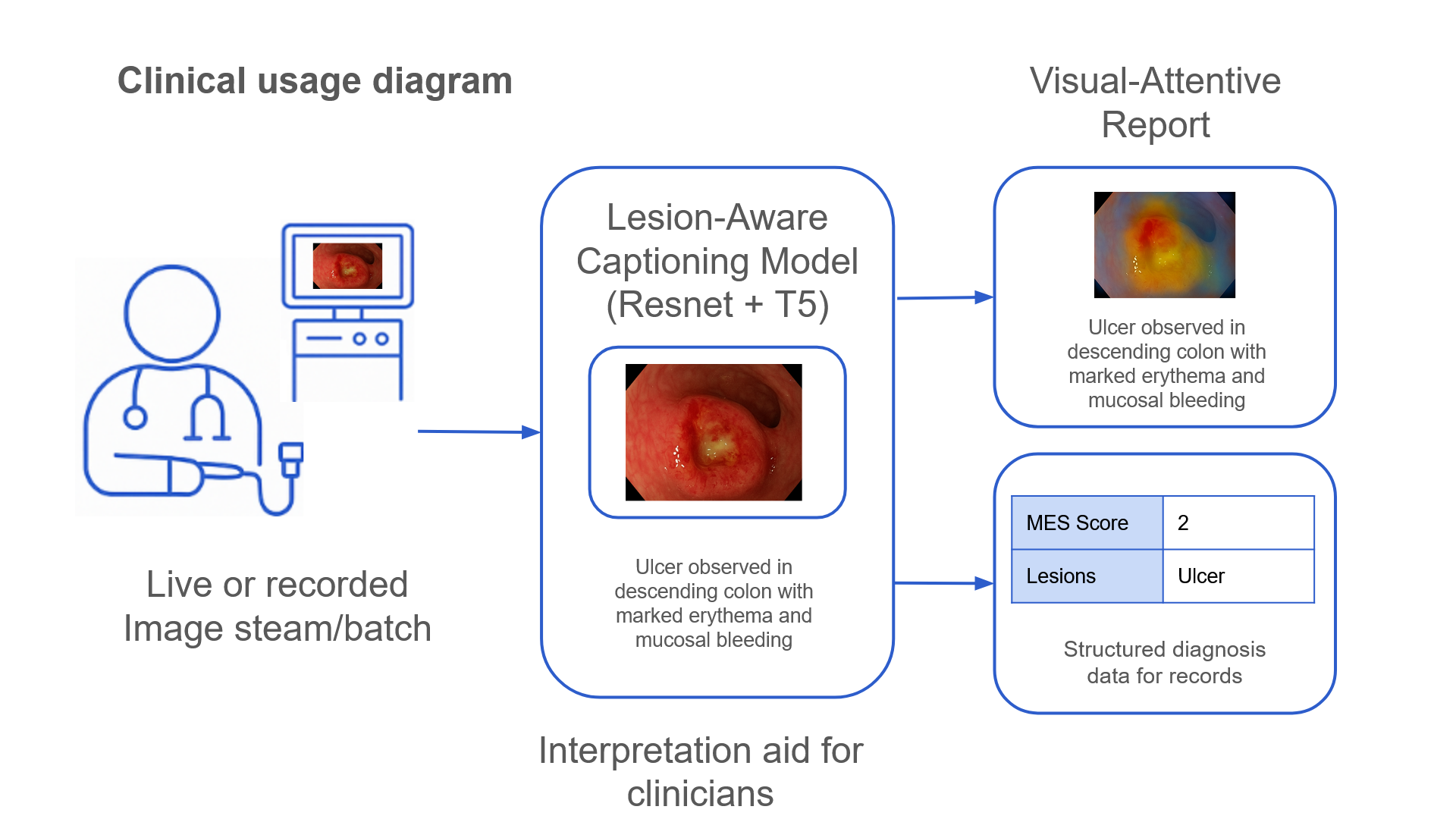} % <-- update with your actual filename and path
    \caption{
        \textbf{Clinical Workflow for Lesion-Aware Image Captioning.}
        This diagram illustrates the clinical workflow for integrating the lesion-aware image captioning model into routine practice. Colonoscopy images are processed by a ResNet + T5-based AI model that generates descriptive captions and structured diagnostic data. The output includes a visual-attentive report with lesion-focused heatmaps and a structured summary of findings, such as MES score and lesion type. These results can assist clinicians in real-time interpretation, reporting, and integration with electronic medical records.
    }
    \label{fig:clinical-workflow}
\end{figure}

This model balances interpretability and diagnostic precision, producing attention-guided, clinically aligned descriptions. Figure~\ref{fig:clinical-workflow} outlines the clinical use case, and Figure~\ref{fig:architecture} illustrates the architectural components.

The remainder of this paper is organized as follows: Section 2, reviews related work. Section 3, details the proposed methodology. Section 4, presents results and ablation. Section 5, concludes with insights and future work.

\section{Theoretical Framework}

Automated image captioning in medical imaging bridges computer vision, natural language processing, and diagnostic knowledge. Early models like \textit{Show and Tell}~\cite{vinyals2015show} and \textit{Show, Attend and Tell}~\cite{xu2015show} introduced CNN-LSTM structures and attention mechanisms to enhance report generation and interpretability, but lacked clinical specificity.

In healthcare, Shin~\textit{et al.}~\cite{shin2016learning} and Jing~\textit{et al.}~\cite{jing2019show} developed domain-adapted captioning methods in radiology, combining hierarchical modeling and visual attention. The shift toward transformers, marked by the introduction of T5~\cite{raffel2020exploring}, enabled scalable, flexible architectures suited to multimodal applications.

In gastroenterology, Valencia~\textit{et al.}~\cite{valencia2023captioning} explored UC captioning using vision-language models, demonstrating how integrating MES scoring and captioning enhances clinical utility. Grad-CAM~\cite{selvaraju2017grad}, widely used for explainability, has been adapted to biomedical settings~\cite{huang2020biomedical} for localizing pathological cues. In UC, combining Grad-CAM with lesion-specific classifiers provides both interpretability and task performance improvements.

For object-level semantics, DETR~\cite{carion2020end} introduced set-based transformers for end-to-end detection. Scene graph methods derived from such models allow structured lesion reasoning and contextual captioning. These advances motivate our lesion-aware fusion framework, which combines visual attention, domain prompts, and transformer-based decoding to produce clinically aligned captions.

Recent MES classification approaches in colonoscopy have yielded moderate-to-high accuracy. Jiang~\textit{et al.}~\cite{jiang2022mes} and Lo~\textit{et al.}~\cite{lo2022efficientnet} used CNNs and EfficientNet-B2, respectively, achieving ~83\% accuracy. Bhambhvani~\textit{et al.}~\cite{bhambhvani2021mes} reported 77.2\% using ResNeXt-101 on HyperKvasir. Valencia~\textit{et al.}~\cite{valencia2023captioning} achieved 77.8\% when combining scoring with captioning. Our model surpasses this with 84.7\% accuracy, demonstrating the benefit of integrating lesion-aware modulation and semantic conditioning.
\section{Methodology}

Lesion-level attention is crucial for interpretability in UC imaging. Grad-CAM~\cite{selvaraju2017grad} localizes clinical features like erythema and bleeding, while multilevel versions~\cite{huang2020biomedical} enhance visual-language performance. Integrating Grad-CAM with lesion classifiers improves spatial focus.

\begin{figure}[t]
    \centering
    \includegraphics[width=1\linewidth]{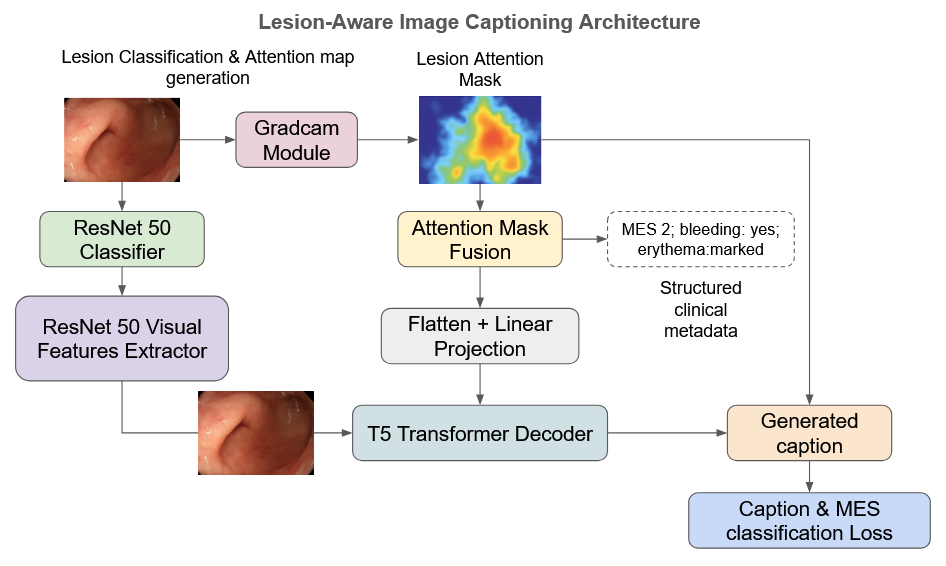} % <-- update with your actual filename and path
    \caption{
        \textbf{Top level architecture for Lesion-Aware Image Captioning.}
        The architecture begins with an RGB endoscopic image processed through two ResNet-50 branches. One branch performs MES classification and generates Grad-CAM attention maps to localize clinically relevant lesions. The second branch extracts visual features, which are refined using a Convolutional Block Attention Module (CBAM) that applies channel and spatial attention. These enhanced features are fused with the Grad-CAM lesion attention map and linearly projected into the embedding space for the T5 transformer decoder. Structured clinical metadata (e.g., MES score, bleeding, erythema) are used as prompts to guide caption generation. The model is trained using a dual loss: cross-entropy for captioning and auxiliary classification loss for MES scoring.
    }
    \label{fig:architecture}
\end{figure}

Transformers like DETR~\cite{carion2020end} enable spatial reasoning via set prediction, supporting structured captioning. Our model (Fig.~\ref{fig:architecture}) fuses CBAM-enhanced ResNet features with Grad-CAM maps and structured prompts, decoded via T5.

We describe the dataset, preprocessing, classification, attention mechanisms, decoding strategy, training procedure, and evaluation protocol in the following subsections.

\subsection{Dataset and Annotation Schema}

The dataset includes 2,355 annotated endoscopic images with captions and UC-specific labels: MES score (0–3), vascular pattern, bleeding, erythema, friability, and ulceration. Labels follow standardized Mayo criteria and were validated by experts. Data was split into training (70\%), validation (15\%), and test (15\%), ensuring MES balance. The dataset was introduced in~\cite{valencia2023captioning}.

\subsection*{Deployment Feasibility and Computational Cost}

Experiments were conducted on an Azure VM with an NVIDIA Tesla T4 GPU (16GB VRAM) and 13GB RAM. Average inference time per image—including Grad-CAM, CBAM, and decoding—was ~1.45 seconds. Memory usage remained under 10GB with batch size 4, supporting real-time or batch deployment. Despite modest hardware, performance remained robust, enabling clinical integration.

\subsection{Lesion Classification and Grad-CAM Generation}

A ResNet-50 was trained for MES classification using cross-entropy loss. Augmentations included flipping, brightness jittering, and cropping. Test accuracy exceeded 84\%. Grad-CAM heatmaps were extracted from the final convolutional layer, upsampled to input size, and normalized. These were used to spatially weight ResNet features.

\subsection{Visual Feature Enhancement via CBAM and Attention Masking}

CBAM and Grad-CAM jointly enhance attention. Grad-CAM heatmaps $M \in \mathbb{R}^{1 \times H \times W}$ are upsampled and normalized. ResNet feature maps $F \in \mathbb{R}^{C \times H \times W}$ are enhanced via:

\[
F' = \text{CBAM}(F) \odot (1 + \alpha M)
\]

Here, $\alpha$ is a learnable scalar. CBAM applies sequential channel and spatial attention. The Grad-CAM mask acts as an external spatial guide, reinforcing lesion relevance.

\subsection{Language Decoder and Cross-Attention Integration}

For the language model, we adopted T5-base due to its success in various text generation tasks. The T5 encoder-decoder was initialized with pretrained weights from Hugging Face Transformers. Visual embeddings $F'$ were flattened and projected into T5-compatible embeddings through a linear transformation followed by layer normalization.

The modified T5 decoder incorporated visual cross-attention at each decoding layer. During caption generation, the decoder used self-attention on previously generated tokens, cross-attention on the visual features, and positional encodings to preserve spatial relationships. We experimented with conditioning the T5 encoder on structured metadata (e.g., \texttt{MES-2; bleeding: yes; friability: moderate}) formatted as a prompt.

\subsection{Training Strategy}

The model is trained end-to-end using:

\[
\mathcal{L}_{\text{total}} = \mathcal{L}_{\text{caption}} + \lambda \mathcal{L}_{\text{MES}}
\]

with $\lambda = 0.2$, selected through validation. Lower values reduced classification accuracy; higher values harmed fluency. This balance optimized semantic richness and diagnostic relevance.

\subsection{Evaluation Metrics}

We report BLEU-4, ROUGE-L, MES accuracy, and alignment score (matching between caption terms and labels). We also evaluate Grad-CAM-to-caption alignment and token-level precision.

\section{Results and Discussion}

\subsection{Quantitative Results}

Our model outperforms the current state of the art in ulcerative colitis image captioning. As shown in Table~\ref{tab:model_comparison}, we achieve \textbf{84.7\%} MES classification accuracy, surpassing the strongest baseline, Valencia et al.~\cite{valencia2023captioning}, which reported \textbf{77.8\%} on the same dataset. This baseline jointly performs captioning and classification over MES 0--3 using a clinically validated dataset. In terms of captioning, our model reaches BLEU-4 \textbf{0.87} and ROUGE-L \textbf{0.85}, improving over Valencia et al.'s \textbf{0.77} and \textbf{0.72}. These gains validate the strength of our lesion-aware design. To assess whether this performance difference is statistically significant, we conducted a paired bootstrap resampling with 1{,}000 iterations, yielding a p-value $<$ 0.01 for both classification and captioning improvements over the baseline.

Key to these results is our attention strategy. Grad-CAM offers spatial supervision by highlighting pathologic regions (e.g., ulcers, bleeding), grounding visual features. CBAM refines features via channel and spatial attention, enhancing focus and suppressing noise. Together, they improve discrimination and vision-language alignment.

Our dual-branch ResNet decouples visual features from MES scoring, reducing interference. Structured clinical prompts (e.g., ``MES-2; bleeding: yes'') guide T5 toward diagnostic phrasing. A language refinement stage improves grammar and clinical terminology for deployment readiness.

To our knowledge, this is the first UC captioning framework to combine: (1) Grad-CAM-based supervision during training, (2) CBAM-based enhancement, and (3) clinical prompt conditioning. While effective individually, their unified integration represents a novel contribution to gastrointestinal image captioning.

\subsection{Qualitative Results}

\begin{figure}[ht!]
    \centering
    \includegraphics[width=0.95\linewidth]{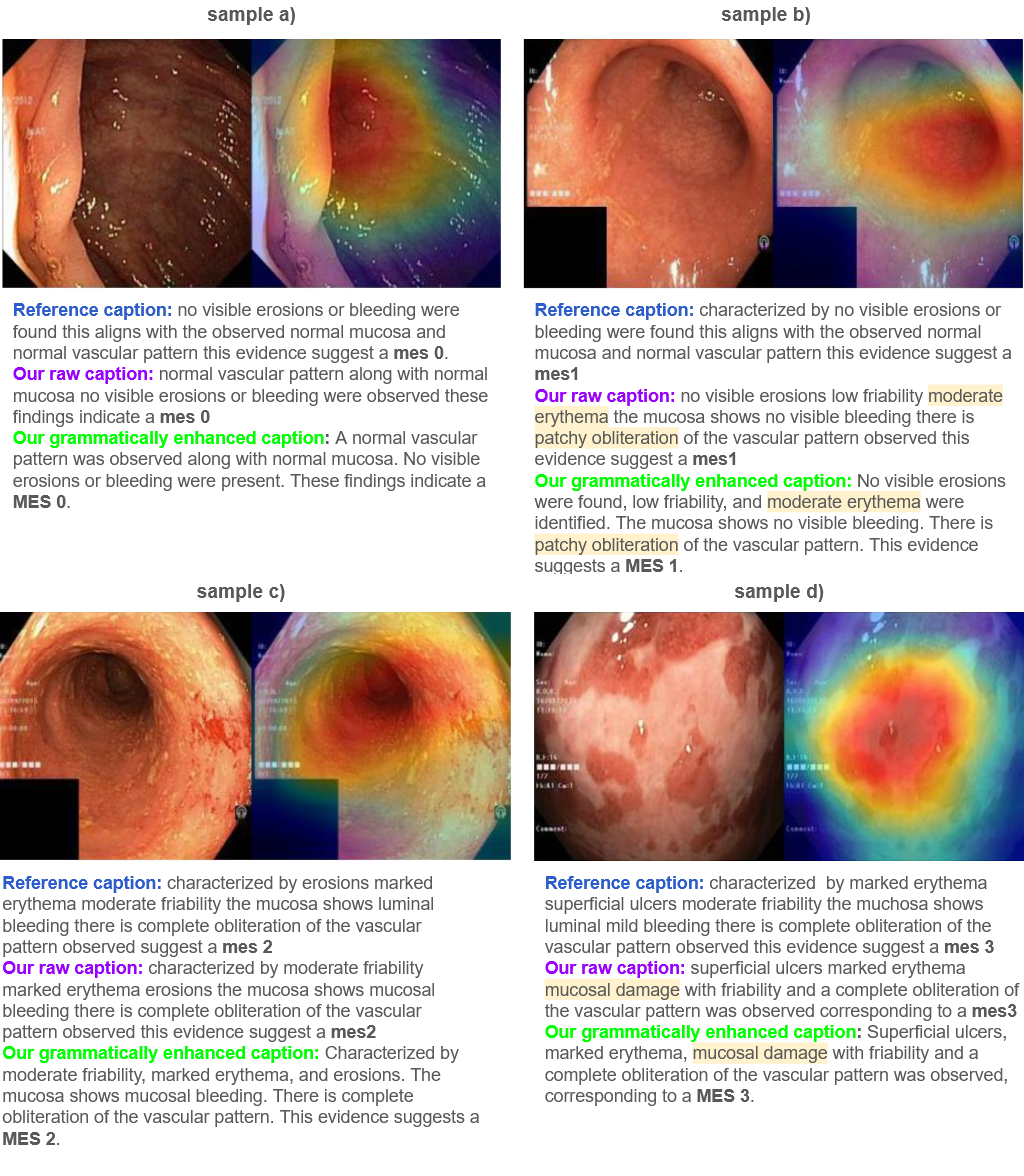} % <-- update with your actual filename and path
        \caption{
        \textbf{Qualitative examples of lesion-aware caption generation across MES grades. }Each panel shows an original endoscopic image (left) and its Grad-CAM overlay (right), highlighting regions that influenced the model’s attention. Below each image, we present the expert caption, the raw model output, and a refined version generated by the T5 correction module. The panels (a–d) represent increasing MES scores from 0 to 3, showcasing spatial focus and linguistic progression across severity levels.
    }

    \label{fig:gradcam_examples}
\end{figure}

To complement the quantitative evaluation, we present a qualitative analysis of our lesion-aware captioning pipeline, illustrated in Figure~\ref{fig:gradcam_examples}. The figure contains four panels (a–d), each representing a different \textbf{MES score} (0 to 3). Within each panel, the \textit{original endoscopic image} is shown on the left, and the \textit{Grad-CAM heatmap overlay} is shown on the right, revealing the model's spatial focus on clinically relevant regions. Each panel includes three textual outputs: the \textit{reference clinical caption}, the \textit{raw caption} generated by our model, and a \textit{grammatically enhanced version} refined using a T5-based post-processor to improve fluency and alignment with clinical language.

In \textbf{panel a} (MES 0), the model accurately identifies a \textit{normal mucosal surface}, with Grad-CAM activations confirming attention to non-pathological regions. The enhanced caption provides a clearer and more formal expression of the normal findings, in line with the reference.

In \textbf{panel b} (MES 1), the model successfully detects features such as \textit{low friability} and \textit{moderate erythema}, which are highlighted in \textit{yellow} in the enhanced caption but are \textit{absent from the reference}. These elements are clinically relevant and consistent with mild inflammation. Grad-CAM activations correctly highlight affected areas, demonstrating effective spatial localization and deeper semantic understanding.

In \textbf{panel c} (MES 2), all captions converge on key pathological signs including \textit{marked erythema}, \textit{friability}, and \textit{complete obliteration} of the vascular pattern. Grad-CAM heatmaps are centered on inflamed mucosa, and the enhanced caption refines structure and tone to improve interpretability.

In \textbf{panel d} (MES 3), the model identifies \textit{superficial ulcers}, \textit{mucosal damage}, and \textit{obliteration of the vascular pattern}, all of which are localized accurately by the Grad-CAM. Notably, some of these descriptors—\colorbox{yellow}{highlighted in yellow}—are present \textit{only in the generated captions} and \textit{missing in the reference}, demonstrating the model's ability to extract nuanced clinical features beyond those originally annotated.

These qualitative results underscore the strength of our proposed architecture—\textbf{ResNet + CBAM + Grad-CAM + T5}—in generating \textit{clinically meaningful}, \textit{interpretable}, and \textit{human-aligned reports}. The \textbf{ResNet backbone} provides robust feature extraction, while the \textbf{CBAM module} enhances sensitivity to relevant visual features via dynamic attention. \textbf{Grad-CAM} further reinforces interpretability by visualizing class-relevant regions, and the \textbf{T5 decoder}, augmented with grammar-aware refinement and metadata prompting, enables fluent, accurate, and domain-aligned language generation.

Together, these components empower the model to identify \textit{subtle disease markers} that may be overlooked in standard reports, as evidenced in panels~\textbf{b} and~\textbf{d} of Figure~\ref{fig:gradcam_examples}. This fusion of visual precision and linguistic fluency not only improves clinical performance but also sets a foundation for \textbf{trustworthy, explainable AI} in gastrointestinal diagnosis.

%TODO: remove green colors - they are very hard to read!!!

\subsection{Ablation Study}

\begin{table}[t]
\centering
\caption{Comparison with the baseline from Valencia et al.~\cite{valencia2023captioning}, showing higher MES classification and captioning performance on the same dataset. The table also includes an ablation study, demonstrating how removing key components like CBAM, Grad-CAM, or clinical prompts reduces overall performance, highlighting their individual contributions.}
\label{tab:model_comparison}
\begin{tabular}{lccc}
\toprule
\textbf{Model} & \textbf{MES Accuracy (\%)} & \textbf{BLEU-4} & \textbf{ROUGE-L} \\
\midrule
Valencia et al. (2023) & 77.8 & 0.77 & 0.72 \\
\textbf{Ours (Lesion-Aware + CBAM)} & \textbf{84.7} & \textbf{0.87} & \textbf{0.85} \\
\midrule
\textit{Ablation Variants} & & & \\
\quad w/o CBAM & 81.2 & 0.83 & 0.79 \\
\quad w/o Grad-CAM & 80.5 & 0.81 & 0.78 \\
\quad w/o Clinical Prompts & 82.1 & 0.82 & 0.80 \\
\quad No Attention Fusion & 78.6 & 0.76 & 0.72 \\
\quad ResNet18 w/o CBAM & 76.4 & 0.74 & 0.70 \\
\bottomrule
\end{tabular}
\label{tab:ablation}
\end{table}
To quantify the contribution of each architectural component, we conducted a set of ablation experiments on our full model (\textbf{ResNet + CBAM + Grad-CAM + T5 with clinical prompts}), which achieves \textbf{84.7\%} accuracy in MES classification, a \textbf{BLEU-4} score of \textbf{0.87}, and a \textbf{ROUGE-L} score of \textbf{0.85}.

First, removing the \textbf{CBAM attention module} resulted in a significant performance drop (MES accuracy: 81.2\%, BLEU-4: 0.83, ROUGE-L: 0.79), confirming the importance of spatial and channel-wise attention for highlighting lesion-relevant features. Similarly, eliminating the \textbf{Grad-CAM integration during training and inference} reduced both interpretability and performance (accuracy: 80.5\%, BLEU-4: 0.81), demonstrating the value of lesion-aware spatial supervision.

Excluding \textbf{clinical prompts} from the T5 decoder led to less precise and semantically aligned descriptions (BLEU-4: 0.82, ROUGE-L: 0.80), reinforcing the role of structured semantic conditioning in language generation. A simplified variant using \textbf{plain feature fusion without any attention mechanism} further degraded results (accuracy: 78.6\%, BLEU-4: 0.76), indicating that the integration strategy between vision and language plays a critical role.

Lastly, switching to a \textbf{lightweight ResNet18 backbone without CBAM} yielded the lowest overall scores (accuracy: 76.4\%, BLEU-4: 0.74), highlighting the need for both a strong visual encoder and attention-guided modulation for effective captioning in medical imaging. These findings validate our architectural choices and emphasize the importance of joint spatial, semantic, and clinical attention in achieving state-of-the-art performance.

Configurations lacking Grad-CAM or CBAM exhibited degraded performance due to insufficient spatial attention to pathological regions. The removal of clinical prompts led to less structured and less specific descriptions, confirming that metadata provides valuable semantic anchors during generation. Notably, the variant without any attention fusion performed similarly to the baseline, indicating that lesion-aware modulation plays a critical role in outperforming prior methods and aligning visual features with clinical language.

\section{Conclusions and future work}

We proposed a lesion-aware captioning framework for ulcerative colitis, combining Grad-CAM, CBAM, and clinical prompts within a ResNet-T5 pipeline. The model outperformed prior work in captioning and MES classification, generating structured and interpretable outputs. Key components—attention fusion and metadata injection—were critical to its success.

Future directions include extending the approach to video colonoscopies, exploring real-time deployment, and incorporating feedback loops for continuous model refinement and clinician trust.

 \section{Acknowledgments}
 The authors acknowledge the support of the “Secretaría de Ciencia, Humanidades, Tecnología e Innovación” (SECIHTI), the French Embassy in Mexico, Campus France, and the Data Science Hub at Tecnológico de Monterrey. This work was partially funded by Microsoft’s AI for Health program (Azure Sponsorship credits), the French-Mexican Ecos Nord grant (MX 322537/FR M022M01), and the Google Explore CSR Program through the LATAM Undergraduate Research initiative.

\bibliographystyle{splncs04}
\bibliography{references}

\begin{thebibliography}{10}
\providecommand{\url}[1]{\texttt{#1}}
\providecommand{\urlprefix}{URL }
\providecommand{\doi}[1]{https://doi.org/#1}

\bibitem{abacha2019vqa}
Abacha, A.B., Shivade, C., Hasan, S.A., Datla, V., Liu, J.D., Demner-Fushman, D.: Vqa-med: Overview of the medical visual question answering task at imageclef 2019. CEUR Workshop Proceedings  (2019)

\bibitem{bhambhvani2021mes}
Bhambhvani, H., Zamora, A.: Deep learning model for automated mayo endoscopic subscore classification. European Journal of Gastroenterology \& Hepatology  (2021)

\bibitem{boag2020baselines}
Boag, W., Lou, J., Zech, J.R., Hughes, M.C., McDermott, M.B.A., Ghassemi, M.: Baselines for chest x-ray report generation. Proceedings of the Machine Learning for Healthcare Conference pp. 371--385 (2020)

\bibitem{carion2020end}
Carion, N., Massa, F., Synnaeve, G., Usunier, N., Kirillov, A., Zagoruyko, S.: End-to-end object detection with transformers. In: European Conference on Computer Vision (ECCV) (2020)

\bibitem{chen2020generating}
Chen, X., Zhang, Z., Kalpathy-Cramer, J., Xing, E.: Generating radiology reports via memory-driven transformer. Medical Image Analysis  \textbf{65},  101786 (2020)

\bibitem{huang2020biomedical}
Huang, X., Wang, R., Xu, L., Gao, Z.: Biomedical image captioning with multi-level attention. Neurocomputing  \textbf{378},  303--312 (2020)

\bibitem{jiang2022mes}
Jiang, X.e.a.: Deep learning-based endoscopic assessment of ulcerative colitis severity. Journal of Gastroenterology  (2022)

\bibitem{jing2019show}
Jing, B., Xie, P., Xing, E.: Show, describe and conclude: On exploiting the structure information of chest x-ray reports. In: Proceedings of the AAAI Conference on Artificial Intelligence (2019)

\bibitem{li2022medical}
Li, X., Zhang, Z., Huang, X., Zhao, Q., Metaxas, D.N.: Medical visual question answering via modular co-attention learning. Medical Image Analysis  \textbf{73},  102196 (2022)

\bibitem{liu2019clinically}
Liu, F., Jin, D., Liu, T., Yu, H.: Clinically accurate chest x-ray report generation. arXiv preprint arXiv:1904.02633  (2019)

\bibitem{lo2022efficientnet}
Lo, C.e.a.: Automatic grading of ulcerative colitis from endoscopic images using efficientnet. Medical Image Analysis  (2022)

\bibitem{raffel2020exploring}
Raffel, C., Shazeer, N., Roberts, A., Lee, K., Narang, S., Matena, M., Zhou, Y., Li, W., Liu, P.J.: Exploring the limits of transfer learning with a unified text-to-text transformer. Journal of Machine Learning Research  \textbf{21}(140),  1--67 (2020)

\bibitem{selvaraju2017grad}
Selvaraju, R.R., Cogswell, M., Das, A., Vedantam, R., Parikh, D., Batra, D.: Grad-cam: Visual explanations from deep networks via gradient-based localization. In: Proceedings of the IEEE International Conference on Computer Vision (ICCV) (2017)

\bibitem{shin2016learning}
Shin, H.C., Roberts, K., Lu, L., Demner-Fushman, D., Yao, J., Summers, R.M.: Learning to read chest x-rays: Recurrent neural cascade model for automated image annotation (2016)

\bibitem{valencia2023captioning}
Valencia, H.e.a.: Lesion-aware image captioning and mes prediction in ulcerative colitis. In: Proceedings of the MICCAI (2023)

\bibitem{vinyals2015show}
Vinyals, O., Toshev, A., Bengio, S., Erhan, D.: Show and tell: A neural image caption generator. In: Proceedings of the IEEE Conference on Computer Vision and Pattern Recognition (CVPR) (2015)

\bibitem{xu2015show}
Xu, K., Ba, J., Kiros, R., Cho, K., Courville, A., Salakhutdinov, R., Zemel, R., Bengio, Y.: Show, attend and tell: Neural image caption generation with visual attention. In: International Conference on Machine Learning (ICML) (2015)

\bibitem{yuan2022hrformer}
Yuan, Y., Chen, X., Wang, J.: Hrformer: High-resolution vision transformer for dense predictive tasks. Advances in Neural Information Processing Systems  \textbf{35} (2022)

\bibitem{zhang2021hierarchical}
Zhang, Z., Liu, T., Chen, Y.: Hierarchical medical image understanding through multi-scale and multi-modal transformer networks. In: Medical Image Computing and Computer-Assisted Intervention (MICCAI) (2021)

\end{thebibliography}
\end{document}